%% file: main.tex
\documentclass[10pt,twocolumn,letterpaper]{article}

\usepackage[pagenumbers]{cvpr} 

\input{preamble}

\definecolor{cvprblue}{rgb}{0.21,0.49,0.74}
\usepackage[pagebackref,breaklinks,colorlinks,allcolors=cvprblue]{hyperref}

\def\paperID{582} 
\def\confName{CVPR}
\def\confYear{2025}

\begin{document}

\title{AudCast: Audio-Driven Human Video Generation \\ by Cascaded Diffusion Transformers}

\author{Jiazhi Guan$^{1,2}$ \quad
Kaisiyuan Wang$^2$ \quad
Zhiliang Xu$^2$ \quad
Quanwei Yang$^5$ \quad
Yasheng Sun$^6$ \quad
Shengyi He$^2$ \\
Borong Liang$^2$ \quad
Yukang Cao$^3$\quad
Yingying Li$^2$\quad
Haocheng Feng$^2$ \quad
Errui Ding$^2$ \quad
Jingdong Wang$^2$ \\
Youjian Zhao$^{1,4\dag}$ \quad
Hang Zhou$^{2\dag}$ \quad
Ziwei Liu$^{3\dag}$ \\
$^1$DCST, Tsinghua University \quad
$^2$Baidu Inc. \quad
$^3$S-Lab, Nanyang Technological University \\
$^4$Zhongguancun Laboratory \quad
$^5$University of Science and Technology of China \quad
$^6$KAUST \\
{\tt\small
{guanjz20@mails.tsinghua.edu.cn, zhouhang09@baidu.com}
}
}

\twocolumn[{
\renewcommand\twocolumn[1][]{#1}%
\maketitle
\vspace{-25pt}
\begin{center}
 \centering
 \includegraphics[width=\textwidth]{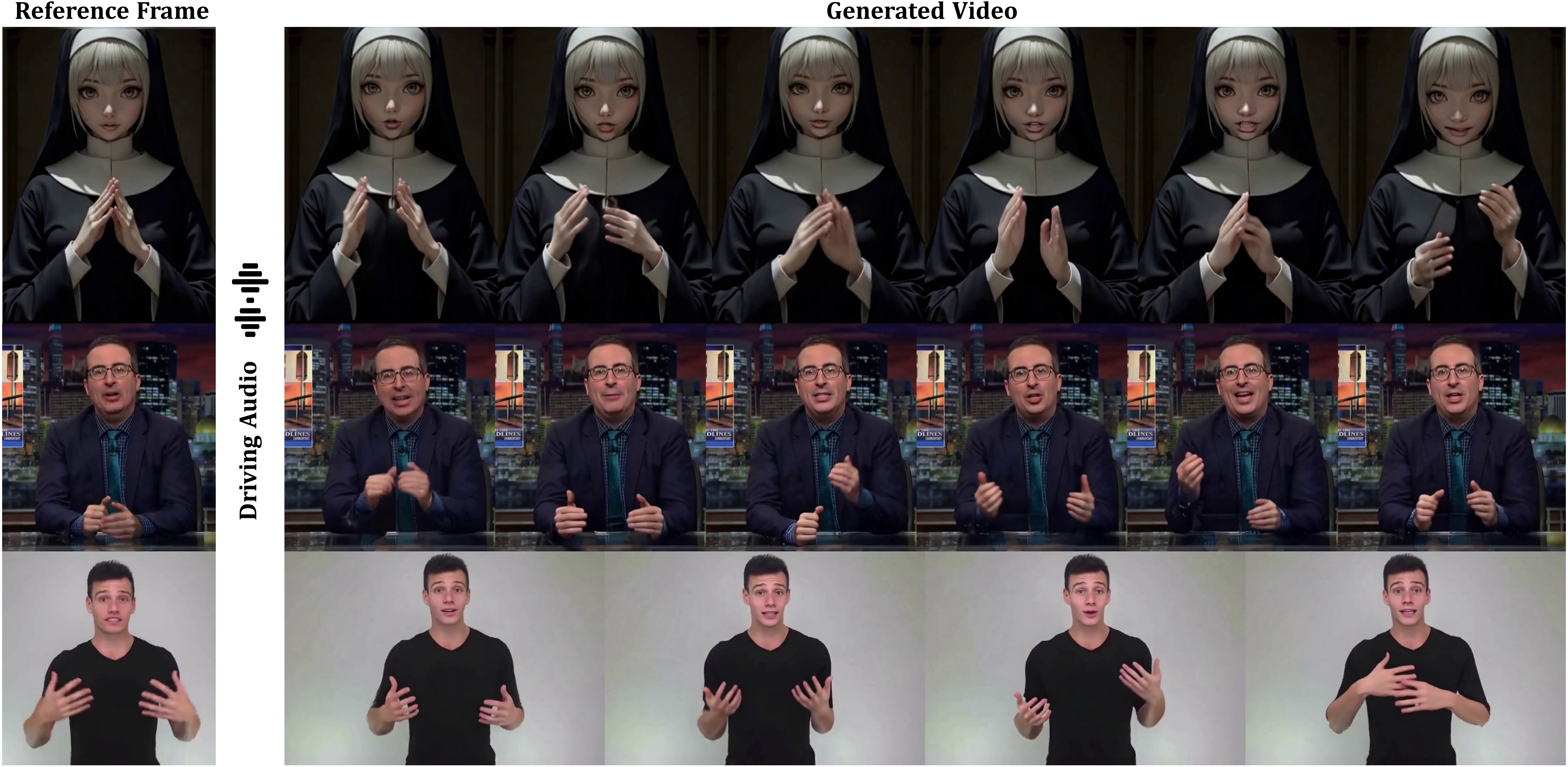}
\vspace{-15pt}
\captionof{figure}{
\textbf{Zero-Shot Results by AudCast.} 
Our method generates lifelike human videos with a realistic style, conditioned on any reference subject and driving audio, in various resolutions.
The synthesized videos exhibit natural, rhythmic motion and expressive expressions, with fine details in both face and hands.
}
\label{fig:teaser}
\end{center}
}]

\maketitle
\def\thefootnote{\dag}\footnotetext{Corresponding Authors.}

\input{sec/0_abstract}    
\input{sec/1_intro}

\input{sec/2_related}
\input{sec/3_method}

\input{sec/4_exp}

\input{sec/5_conclusion}

{
    \small
    \bibliographystyle{ieeenat_fullname}
    \bibliography{main}
}

\end{document}

%% file: preamble.tex
\usepackage{graphicx}
\usepackage{amsmath}
\usepackage{amssymb}
\usepackage{booktabs}
\usepackage{color}
\usepackage{multirow}
\usepackage[dvipsnames]{xcolor}
\usepackage{colortbl}
\usepackage{arydshln}
\usepackage{tabularx} 
\usepackage[accsupp]{axessibility}  

%% file: sec/0_abstract.tex
\begin{abstract}
Despite the recent progress of audio-driven video generation, existing methods mostly focus on driving facial movements, leading to non-coherent head and body dynamics. 
Moving forward, it is desirable yet challenging to generate holistic human videos with both accurate lip-sync and delicate co-speech gestures \wrt given audio.
In this work, we propose \textbf{AudCast}, a generalized audio-driven human video generation framework adopting a cascade Diffusion-Transformers (DiTs) paradigm, which synthesizes holistic human videos based on a reference image and a given audio. \textbf{1)} Firstly, an audio-conditioned \textbf{Holistic Human DiT} architecture is proposed to directly drive the movements of any human body with vivid gesture dynamics. \textbf{2)} Then to enhance hand and face details that are well-knownly difficult to handle, a \textbf{Regional Refinement DiT} leverages regional 3D fitting as the bridge to reform the signals, producing the final results.  Extensive experiments demonstrate that our framework generates high-fidelity audio-driven holistic human videos with temporal coherence and fine facial and hand details. 
Resources can be found at \url{https://guanjz20.github.io/projects/AudCast}.

\end{abstract}

%% file: sec/1_intro.tex
\section{Introduction}
\label{sec:intro}
Nowadays, audio-driven virtual humans have played important roles in various scenarios, serving as virtual hosts, lecturers, and salespeople. Particularly, alongside the usages of large language models~\cite{touvron2023llamaopenefficientfoundation,qwen}, virtual humans are crucial in boosting human-computer interaction experiences by providing interactive visual appearances.
While great efforts have been paid to the field of audio-driven human animation~\cite{prajwal2020lip,zhou2019talking,ji2021audio,zhou2021pose,guan2023stylesync,guan2024resyncer,xu2024hallo}, most previous research focus on controlling only the lip or torso areas~\cite{zhang2024personatalk,vpgc,vprq}. 
Though lip-sync methods aim to blend lip movements into videos seamlessly, their body movements cannot be modified, leading to inconsistent body rhythm and speech. 
Moreover, showing only talking heads cannot satisfy the needs of most scenes. 
It is essential to produce audio-coherent avatars
with accurate lip-sync and rhythmic gestures.

In order to generate real-world human videos with co-speech gestures, a few studies~\cite{liu2022audio,he2024co} map audios to implicit motion representations~\cite{siarohin2021motion} in warpping-based generative models. However, such models cannot produce high-resolution results with precise details.
Meanwhile, researchers have been exploring speech-driven gesture generation~\cite{liu2022learning,zhi2023livelyspeaker,yang2023diffusestylegesture,yi2022generating,zhu2023taming,chen2024diffsheg} in the representation of explicit 3D key points and body meshes. However, the data in these works relies on 3D key point detection and 3D reconstruction~\cite{MANO:SIGGRAPHASIA:2017,SMPL-X:2019}, which could easily be inaccurate due to the ill-pose nature of single-view 3D reconstruction. 
The limited availability of high-quality captured data restricts the broader applicability of these methods in general settings.
As a result, despite the recent progress in 2D- and 3D-based conditional human video generation~\cite{hu2023animate,huang2024make,chang2023magicpose,peng2024controlnext} with diffusion models~\cite{ho2020denoising,song2020denoising,blattmann2023stable}, only few studies have made the exploration of leveraging explicit audio-driven 2D and 3D intermediate representations to recover real-world humans. Particularly, Vlogger~\cite{corona2024vlogger} proposes a two-stage framework with diffusion models and generates videos in a coarse-to-fine manner, but the movements are not natural enough.

Inspired by recent success in text-based video generation~\cite{yang2024cogvideox}, that multi-modal condition enables high-quality generations with large-scale training,
an ideal solution is to directly learn the mapping between audio and human videos, which is particularly challenging but promising. 
To facilitate this goal, two particular questions need to be answered:
\textbf{1)} How to build a powerful multi-modal generative backbone that suits audio-driven human video generation? 
\textbf{2)} It is well-known that generating human animations, particularly intricate elements such as finger gestures, remians a challenging task even in image synthesis. How do we preserve the fine details of holistic human appearance?

In this work, we propose \textbf{AudCast}, the framework for \textbf{Au}dio-\textbf{D}riven human video generation by \textbf{Cas}caded Diffusion \textbf{T}ransformers with local refinement, which holistically produces expressive and realistic audio-driven human dynamics from photos. 
Our key is to \emph{take advantage of both holistic audio-driven body movement generation and regional detail refinements with Diffusion Transformers (DiTs)}. 
Specifically, we build our model upon the recent success in text-to-video (t2v) generation, where DiT~\cite{peebles2023scalable} has been proven effective in generating high-quality images~\cite{esser2024scaling} and videos~\cite{opensora,yang2024cogvideox} that are coherent with multi-modality conditions. Thus, we first propose the \emph{Holistic Human DiT (H$^2$-DiT)}
upon an existing t2v backbone and make concise yet principal modifications by involving the conditions of audio, appearance, and other functional inputs,
facilitating effective audio-driven animation that consistently preserves the target's appearance and produces audio-coherent movements.

However, even large-scale video generation models cannot handle delicate human details such as fingers. Similarly, our H$^2$-DiT might produce blurry results on details, particularly when the predicted gestures are complicated. 
We observe that while the generated results may lack precision in complex regions such as hands and faces, they can be effectively refined using localized 3D reconstruction methods~\cite{pavlakos2024reconstructing,deng2019accurate}, which provide rich structural information.
Rather than explicitly utilizing intermediate 3D representations as direct guidance for whole-body movement generation~\cite{huang2024make,zhu2024champ}, which can constrain the natural dynamics of body motion, we instead leverage these structural priors solely for localized refinement.
The rendered 3D parts offer explicit guidance by supplying sufficient structural information, which benefits more fine-grained regional synthesis on the corresponding local areas.
We thereby modify the DiT structure to a \textit{Regional-Refinement DiT (R$^2$-DiT)} that takes the 3D guidance to produce more realistic results with intricate details. Experiments show that the results from our cascaded DiTs clearly outperform previous studies.

Our contributions can be summarized as follows:
\textbf{1)} We propose the \textbf{AudCast} framework to tackle the task of audio-driven human video generation, which produces expressive co-speech human videos with intricate details. 
\textbf{2)} 
With straightforward yet effective design choices, we develop the Holistic Human DiT, which integrates multi-modal conditions and key foundational components to enable generalized, long-duration human video generation with rhythmic motion and expressive expressions.
\textbf{3)} 
Leveraging local 3D reconstructions, we propose the Regional Refinement DiT, a model that enhances the outputs of the holistic video generation through a cascaded process informed by 3D structural priors. This refinement yields noticeable improvements in fine details of the face and hands.

%% file: sec/2_related.tex
\section{Related Works}
\label{sec:related}

\noindent\textbf{Audio-Driven Human Gesture Generation}.
A body of research focuses on generating human animations by modeling the mapping relationship between audio signals and human skeletons. Previous studies have explored generative models such as GANs~\cite{goodfellow2014generative}, VAEs~\cite{kingma2013auto}, etc. More recently, advancements in diffusion models have led to a series of studies that leverage these models for more refined and realistic human motion synthesis~\cite{yang2023diffusestylegesture,zhu2023taming,chen2024diffsheg,chhatre2024emotional}. 
Several studies~\cite{he2024co,liu2022audio} bypass structural representations, generating gesture images directly from audio inputs. 
Moreover, a concurrent study~\cite{liu2024tango} adopts a retrieval-based solution. It relies on a reference video to drive gesture reenactment.
While effective within constrained settings, these approaches are tailored to specific subjects’ motion patterns and appearance, lacking the generalization required for broad human video generation as we address in this paper.

\noindent\textbf{Pose-Driven Human Video Reenactment}.
Recent studies~\cite{bhunia2023person,liao2024appearance,wang2023disco,zhu2024champ,xu2023magicanimate,hu2023animate,zhu2024champ,chang2023magicpose,peng2024controlnext,zhang2024mimicmotion,guan2024talk,yang2024showmaker} have extensively explored pose-driven human video reenactment using diffusion models based on UNet~\cite{ronneberger2015u}. 
In these works, the driving signal is typically represented as a human skeleton or more complex representations like 3D rendering.
This pose input is either incorporated into the diffusion process by concatenating it with the initial noises or integrated into multi-scale features, often with the aid of a ControlNet~\cite{zhang2023adding}. These methods facilitate detailed control over pose dynamics throughout the diffusion process.

While these methods yield promising results, they are not designed to directly address the challenge of our work, which involves mapping audio signals to human motion in video generation. 
However, by utilizing previously discussed audio-driven human gesture generation models as the first stage, these pose-driven methods can function as a second stage, generating human videos conditioned on the outputs from the first stage.
In our experiments, we evaluate these combined approaches as baselines.

\noindent\textbf{Audio-Driven Human Video Generation}.
Directly generating human videos from audio input is a challenging task and has seen limited exploration. Vlogger~\cite{corona2024vlogger} is among the first to tackle this problem, using a two-stage diffusion-based framework. In this approach, 3D body parameters are estimated in the first stage to serve as driving signals, guiding human motion generation in the second stage. However, intermediate representations often introduce accumulative error, which can degrade the overall quality. This limitation underscores the need for models that can directly interpret audio features into human videos.
To the best of our knowledge, only one concurrent study~\cite{lin2024cyberhost} addresses general human video generation by incorporating specialized cross-attention layers within a UNet architecture. In contrast, our approach leverages sequential modeling of multimodal tokens, enabling us to generate coherent and dynamic gesture patterns directly. 

\begin{figure*}[!t]
\centering
\includegraphics[width=\linewidth]{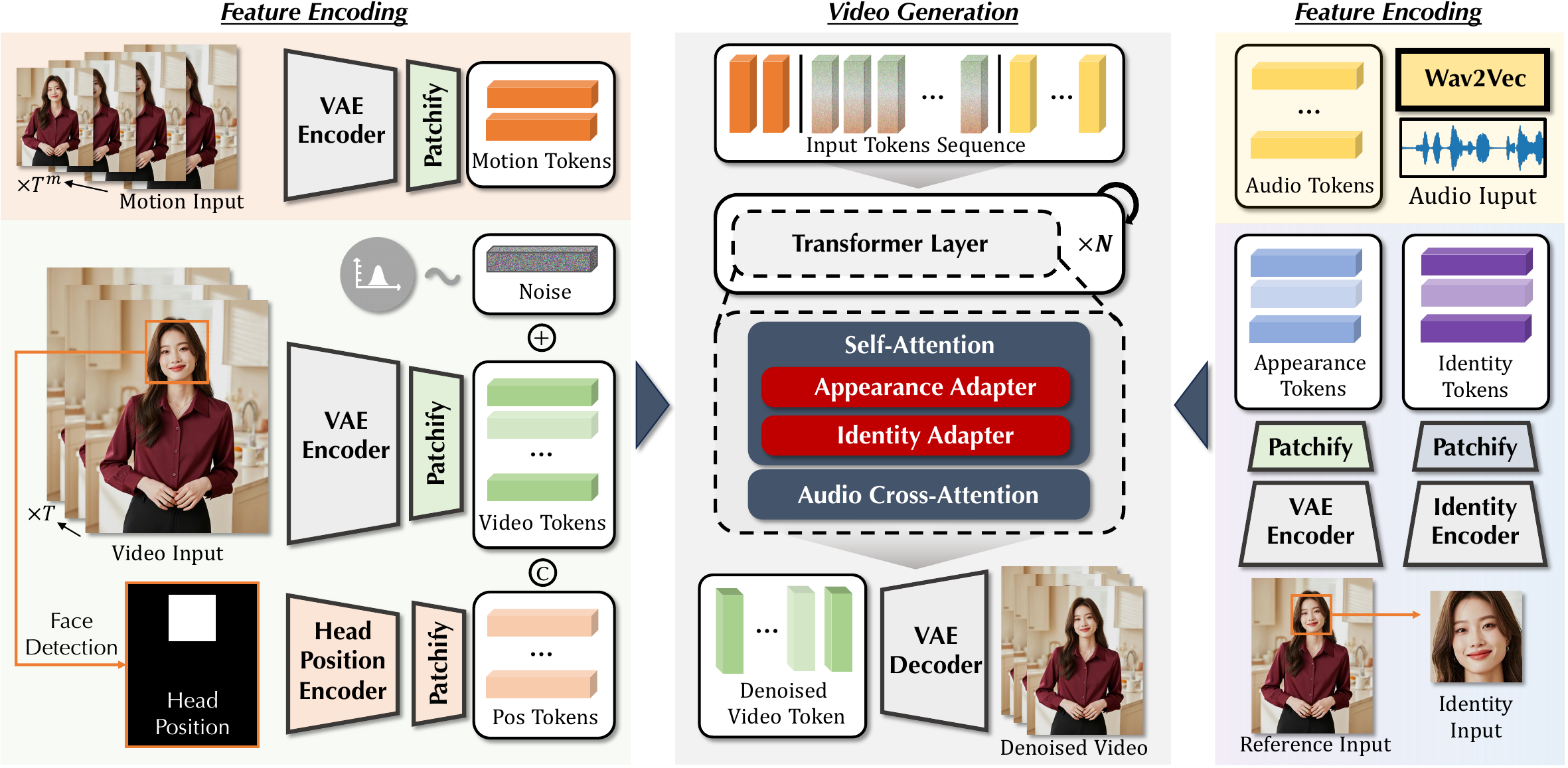}
\caption{
\textbf{Holistic Human DiT}. 
We illustrate the model from two phases including \textit{Feature Encoding} and \textit{Video Generation} in the figure. 
In the first phase, we adopt several modules to extract rich features and tokenize them into sequential inputs.
During generation, our DiT-based model handles these inputs and performs iterative denoising to get the final videos.
}
\vspace{-15pt}
\label{fig:pipeline}
\end{figure*}

\noindent\textbf{Diffusion-Based Video Generation}.
Right now, diffusion-based models are transforming video generation~\cite{bar2024lumiere,blattmann2023stable,he2022latent,ho2022imagen,khachatryan2023text2video}, especially with the framework of VQVAE~\cite{van2017neural} and Transformers~\cite{vaswani2017attention}. We have seen impressive video results produced by models like SORA~\cite{videoworldsimulators2024}, Pika, and Kling.
Our work is greatly inspired by these advancements, particularly the open-sourced CogVideoX~\cite{yang2024cogvideox}.
Despite significant progress, general video generation models often struggle to capture nuanced human motions, especially in detailed areas like the face and hands. 
Our method not only extends human video generation to be audio-driven but also enhances the quality of fine motion generation, providing more realistic and expressive human dynamics.

%% file: sec/3_method.tex
\section{Methodology}
\label{sec:method}

This work tackles the challenge of holistic audio-driven human video generation in a realistic style. Given processed audio features $\textbf{a} = \{a_{t_s}, \cdots, a_{t_e}\}$, which serve as the driving signals, and a single reference frame $I_{ref}$ providing appearance information, our objective is to generate vivid human videos with natural motion that aligns with the driving audio while maintaining visual consistency with the reference frame. Here, $t_s$ and $t_e$ represent the corresponding time stamps.

In this work, we leverage a pretrained 3D Causal VAE~\cite{yang2024cogvideox} for video compression, where the encoder and decoder are represented by $\mathcal{E}$ and $\mathcal{D}$, respectively. 
Given a target video clip $\textbf{V} = \{I_{t_s}, \cdots, I_{t_e}\}$,
our method operates on the compressed visual latent features, represented by 
$\textbf{Z}=\mathcal{E}(\textbf{V})$.
During training, our model is optimized to restore the video latent $\hat{\textbf{Z}}$ from the noised latent $\textbf{Z}'$, which is produced through a predefined diffusion process~\cite{ho2020denoising}.
The restored latent is then decoded to reconstruct the video in pixel space as $\hat{\textbf{V}}=\mathcal{D}(\hat{\textbf{Z}})$.

\subsection{Holistic Human DiT}
An overview of the proposed Holistic Human DiT (H$^2$-DiT) is shown in Fig.~\ref{fig:pipeline}. 
We first give a general explanation of our approach in two phases: feature encoding and video generation.
In the feature encoding phase, we integrate several specialized modules to effectively encode the input features and incorporate supplementary information that ensures consistency across generated frames.
In the video generation phase, we build our model with stacked Transformer layers that process the tokenized features and iteratively perform denoising steps.

Starting with the video latent $\textbf{Z}$, we first divide the features into patches. These patches are then linearly embedded to form a sequence of \textit{video tokens} $\textbf{T}\in \mathbb{R}^{S^v\times E}$, where $S^v$ represents the sequence length, and $E$ denotes the embedding dimension in Transformer layers. 
Subsequently, a randomly sampled noise is added to produce a sequence of noisy video tokens $\textbf{T}'$.

\noindent\textbf{Head Position Guidance}.
Driving human gestures with audio solely as conditions presents a well-known challenging many-to-many mapping problem~\cite{zhu2023taming,chen2024diffsheg}, even within a simplified, sparse parametric 3D representation. While in the context of video generation, the challenge is further compounded by the high degree of freedom and increased complexity.
With a task-tailored consideration that the head movements are typically minimal in human videos, we propose to include a Head Position Encoder,
which consists of several downsample 3D convolutional layers, to provide a weak control as similar designs have shown to be effective in talking head generation~\cite{tian2024emo} and customized image generation~\cite{wang2024instantid,zhang2024flashface}. 
Specifically, we first detect the facial regions and create a mask $\textbf{S}$ covering the head movements across all the frames.
Next, the mask is processed through the Head Position Encoder, followed by a patch embedding layer that splits the features into patches $\textbf{T}^{P}\in \mathbb{R}^{S^v\times E}$. This control information is then integrated with the video tokens by channel concatenation and an MLP layer, denoted as $\textbf{T}^{P'}=\operatorname{MLP}(\operatorname{Cat}^c(\textbf{T}^{P}, \textbf{T}'))$, creating a unified sequence of the video tokens.

\noindent\textbf{Motion Condition}.
Our objective is to generate extended video sequences that continue seamlessly for the full duration of the driving audio.
However, most DiT-based video generation frameworks are limited in this regard due to GPU memory constraints, which restrict their ability to directly support long-duration video generation.
Inspired by talking head generation~\cite{tian2024emo,xu2024hallo} that incorporating additional visual cues from prior frames can enhance temporal coherence, we adopt a similar approach to seam two independent inferences. Specifically, we encode the previous $M$ frames of the video input, $\textbf{M} = \{I_{t_s-M}, \cdots, I_{t_s-1}\}$, to a representation as \textit{motion tokens} $\textbf{T}^M$, using the same VAE and linear patch embedding employed for the video input. These embedded motion tokens are then concatenated with the video tokens along the temporal dimension for the denoising processing. Thus the model learns to predict consistent appearance and movements taking motion tokens as initialized points.

Note that prior frames would be unavailable in the initial forward pass during inference, we address this by applying a dropout ratio of 0.5 to $\textbf{T}^M$ during training. This encourages the model to generalize effectively in their absence, enhancing robustness for inference. 
Additionally, applying a relatively high dropout rate effectively addresses the issue of information leakage. Since appearance information is also preserved in the motion tokens, relying heavily on details from previous frames can lead to significant error accumulation during long-video generation. We experimentally find that a dropout rate of 0.5 substantially alleviates this issue, promoting a more stable appearance restoration from the reference frame, which will be detailed later.

\noindent\textbf{Audio Embedding}.
Following a series of works~\cite{xu2024hallo,guan2024resyncer,Xing_2023_CVPR} in multi-modal animation, a pretrained wav2vec~\cite{baevski2020wav2vec} model should be a robust option for audio feature encoding. We also adopt this model and project the last 12 layers of audio embeddings into sequential \textit{audio tokens} denoted as $\textbf{A}\in \mathbb{R}^{S^a\times E}$. These tokens are subsequently concatenated with the noisy video tokens, supplying rhythmic information that guides the denoising process to generate holistic human movements. 

\noindent\textbf{Transformer Layer}.
We build a single Transformer layer following MM-DiT~\cite{esser2024scaling} but replace the text tokens with our audio tokens. 
One remaining problem is how to effectively conduct denoising following the appearance in the reference input. Inspired by the adapter~\cite{ye2023ip} for conception injection, we introduced two similar modules in the self-attention layer of the MM-DiT block, namely Appearance Adapter and Identity Adapter. Thereby writing the modified self-attention as:
\vspace{-5pt}
\begin{align}
\text{}&\text{Att}_{\textbf{W}^R_k, \textbf{W}^R_v, \textbf{W}^F_k, \textbf{W}^F_v}(Q,K,V, \textbf{R}, \textbf{F}, \bar{\textbf{S}}) = \nonumber \\
& \text{softmax}\left(\frac{Q {K}^\top}{\sqrt{c}}\right) {V}
 + \text{softmax}\left(\frac{Q (\textbf{R}\textbf{W}_k^R)^\top}{\sqrt{c}}\right) \textbf{R}\textbf{W}_v^R \nonumber \\
& + \bar{\textbf{S}} \cdot \text{softmax}\left(\frac{Q (\textbf{F}\textbf{W}_k^F)^\top}{\sqrt{c}}\right) \textbf{F}\textbf{W}_v^F ,
\end{align}
where $\textbf{R}$ and $\textbf{F}$ represent the appearance tokens and identity tokens, respectively. Specifically, $\textbf{R}$ is derived from the reference input, encoded and split into patches following the encoding process of video tokens. Meanwhile, $\textbf{F}$ is obtained by patchifying the features extracted from a pretrained ArcFace~\cite{deng2019arcface};
$\textbf{W}^R_k, \textbf{W}^R_v, \textbf{W}^F_k, \textbf{W}^F_v$ are four matrices introduced to get \textit{Key} and \textit{Value} for $\textbf{R}$ and $\textbf{F}$, and $\bar{\textbf{S}}$ represents a sequential mask derived from the previously applied head mask $\textbf{S}$. It enables the attention to focus specifically on the facial region, allowing it to attend effectively to identity-related features given by $\textbf{F}$.

In addition, audio serves as the sole driving signal with a dual purpose: 1) guiding human gestures and 2) controlling facial expressions, particularly lip movements. While the self-attention layers effectively handle the first aspect, the alignment of facial expressions with audio is critical, as lip dynamics need to precisely match the timing of speech. 
To enhance this alignment, we explicitly build this correspondence by introducing an additional audio cross-attention layer.
This can be learned from:
\vspace{-10pt}
\begin{equation}
\text{Audio-Att}(Q, \textbf{A}, \bar{\textbf{S}}) = \bar{\textbf{S}} \cdot \text{softmax}\left(\frac{Q \textbf{A}^\top}{\sqrt{c}}\right) \textbf{A},
\end{equation}
where $Q$ represents the tokenized features processed during the DiT forward pass, $\bar{\textbf{S}}$ specifies the cross-attention on the head region.

\begin{figure}[!t]
\centering
\includegraphics[width=0.95\linewidth]{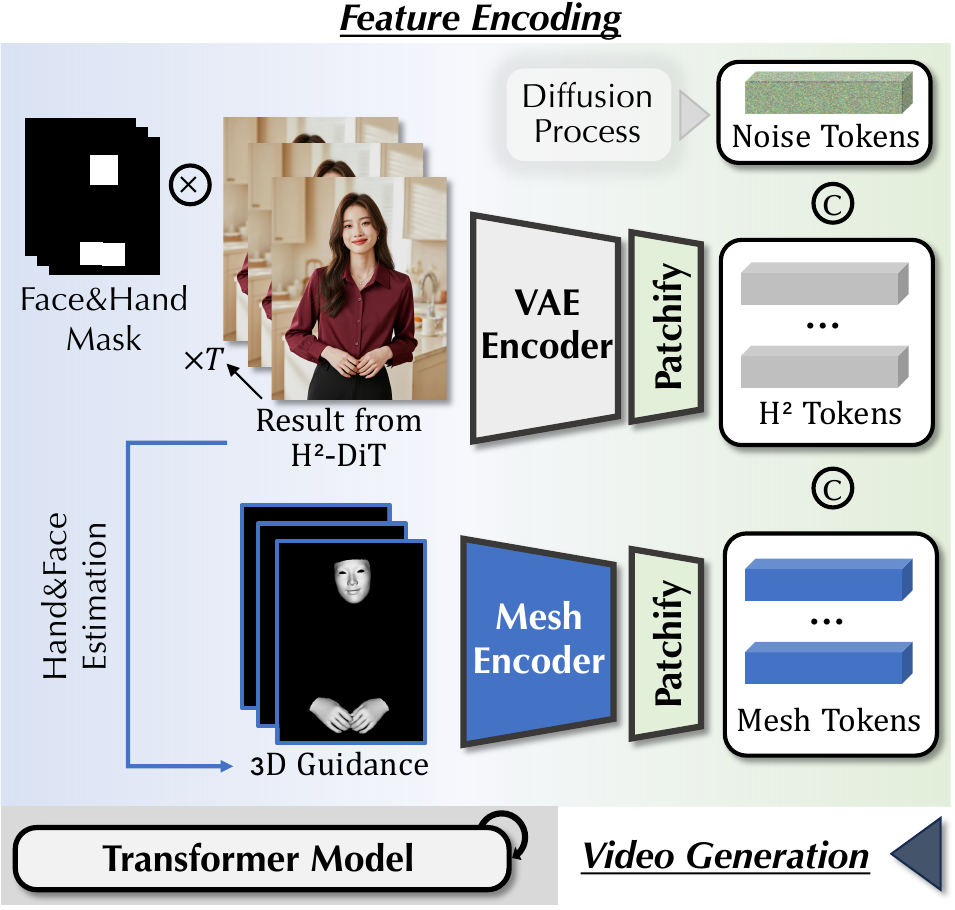}
\vspace{-5pt}
\caption{
\textbf{Regional Refinement DiT}. 
3D structural priors are leveraged to refine the details of the face and hands in a conditional inpainting manner.
}
\vspace{-15pt}
\label{fig:2nd_stage}
\end{figure}

\subsection{Regional Refinement DiT}
Generating fine details of the human face and hands is well-known challenging, which poses an even more critical problem when involving cross-modal generation in our task. 
To enhance details in our audio-driven results, we train a cascaded Regional Refinement DiT (R$^2$-DiT). 
This is inspired by recent studies~\cite{guan2024talk,zhou2024realisdance,lu2024handrefiner} that 3D structural priors can enhance generation quality.

As illustrated in Fig.~\ref{fig:2nd_stage}, we first estimates the 3D representations of both the face and hands using well-established methods~\cite{deng2019accurate,pavlakos2024reconstructing}. These 3D priors are then encoded and tokenized with a Mesh Encoder, which shares the same architecture as the previously introduced Head Position Encoder.
Next, we concatenate the tokenized features (\textit{mesh tokens}) with the representations (\textit{H$^2$ tokens}) extracted from the video generated by H$^2$-DiT, selectively masking the regions corresponding to the face and hands for inpainting. This design enables the model to refine and complete these areas using contextual information.
Subsequently, we concatenate these tokenized features with \textit{noise tokens}, which are generated following the same diffusion process as H$^2$-DiT. The concatenated tokens are then processed by a Transformer-based architecture similar to H$^2$-DiT to synthesize the final video output.
Notably, since human motion has already been synthesized by H$^2$-DiT, the audio signal is no longer required in R$^2$-DiT. Consequently, both the Audio Tokens and Audio Cross-Attention layers are removed from the model.

%% file: sec/4_exp.tex
\begin{table*}[]
\centering
\caption{\textbf{Quantitative Results}. 
Results denoted with * are evaluated on the Vlogger demo set, while those marked with \dag ~pertain specifically to the subject Oliver~\cite{ahuja2020style}. All other results are obtained on a combined test set, which includes our dataset alongside video samples from PATs~\cite{ahuja2020style}.
Ablation results are listed in the bottom rows, please refer to Sec.~\ref{sec:ablation} for the notations.
}
\vspace{-5pt}
{
\resizebox{\linewidth}{!}{
\renewcommand{\arraystretch}{1}
\begin{tabular}{l|ccc|cc|cccc|c}
\toprule
\multirow{2}{*}{Method} & \multicolumn{3}{c|}{\textit{Visual Quality}}       & \multicolumn{2}{c|}{\textit{Temporal Coherence}} & \multicolumn{4}{c|}{\textit{Animation Quality}} & \multicolumn{1}{c}{\textit{Identity Fidelity}} \\
\cline{2-5}
\cline{5-7}
\cline{7-11}
 & SSIM$\uparrow$  & LPIPS$\downarrow$ & FID$\downarrow$ & FID-VID$\downarrow$       & FVD$\downarrow$ & BAS$\uparrow$ & Body-C$\uparrow$ & Hand-C$\uparrow$ & Hand-V${\scriptstyle \times 10^{2}}\uparrow$ & CosSim$\uparrow$ \\ 
\hline
S2D \dag            & 0.7462        & 0.2056        & 67.00         & 20.84         & 743.93        & 0.2258        & 0.9621        & 0.7989        & \textbf{0.6065}        & \textbf{0.8325}        \\
Ours \dag           & \textbf{0.7832}        & \textbf{0.1606}        & \textbf{42.63}         & \textbf{16.25}         & \textbf{504.75}        & \textbf{0.2260}        & \textbf{0.9642}        & \textbf{0.8493}        & 0.2859        & 0.7527        \\
\hline
Vlogger*         & -        & -        & -             & -             & -             & -       & \textbf{0.9231}        & 0.8407        & 0.1855        & 0.6484        \\
Ours*            & -       & -        & -             & -             & -             & -        & 0.9059        & \textbf{0.8522}        & \textbf{0.4037}        & \textbf{0.7302}        \\
\hline
Talk-Mimic      & 0.5497        & 0.4375        & 136.82        & 134.02        & 1648.92       & 0.2257        & 0.8976        & 0.7465        & 0.3467        & 0.4190        \\
Talk-CNxt       & 0.6670        & 0.3020        & 80.49         & 85.44         & 1289.06       & 0.2187        & 0.9364        & 0.7633        & 0.4127        & 0.4887        \\
Prob-Mimic      & 0.5546        & 0.4400        & 131.25        & 112.85        & 1563.95       & 0.2297        & 0.9172        & 0.8691        & 0.4365        & 0.3265        \\
Prob-CNxt       & 0.6630        & 0.2966        & 74.20         & 62.65         & 1420.41       & 0.2260       & 0.9352        & 0.8677        & \textbf{0.4642}        & 0.6160        \\
Ours            & \textbf{0.8275}        & \textbf{0.1368}        & \textbf{45.41}         & \textbf{37.45}         & \textbf{527.24}        & \textbf{0.2355}        & \textbf{0.9485}        & \textbf{0.8832}        & 0.3401        & \textbf{0.7454}        \\
\hline
w/o R$^2$-DiT        & 0.8118        & 0.1399        & 48.27         & 37.57         & 549.01        & 0.2330        & 0.9428        & 0.8579        & 0.3319        & 0.7248        \\
w/o MT          & 0.8023        & 0.1795        & 45.75         & 37.52         & 629.30        & 0.2343        & 0.9424        & 0.8283        & 0.3264        & 0.7376        \\
w/o HPE         & 0.7999        & 0.1416        & 50.37         & 35.13         & 660.13        & 0.2320        & 0.9319        & 0.8526        & 0.3278        & 0.7331        \\
w/o AA          & 0.7253        & 0.1927        & 63.94         & 35.11         & 667.29        & 0.2309        & 0.9457        & 0.8408        & 0.3395        & 0.7016        \\
w/o IA          & 0.7626        & 0.1454        & 48.69         & 37.59         & 575.94        & 0.2320        & 0.9419        & 0.8434        & 0.3376        & 0.6839        \\
\bottomrule
\end{tabular}
}
}
\vspace{-10pt}
\label{tab:cmp_sota_metrics}
\end{table*}

\begin{figure*}[]
\centering
\includegraphics[width=\linewidth]{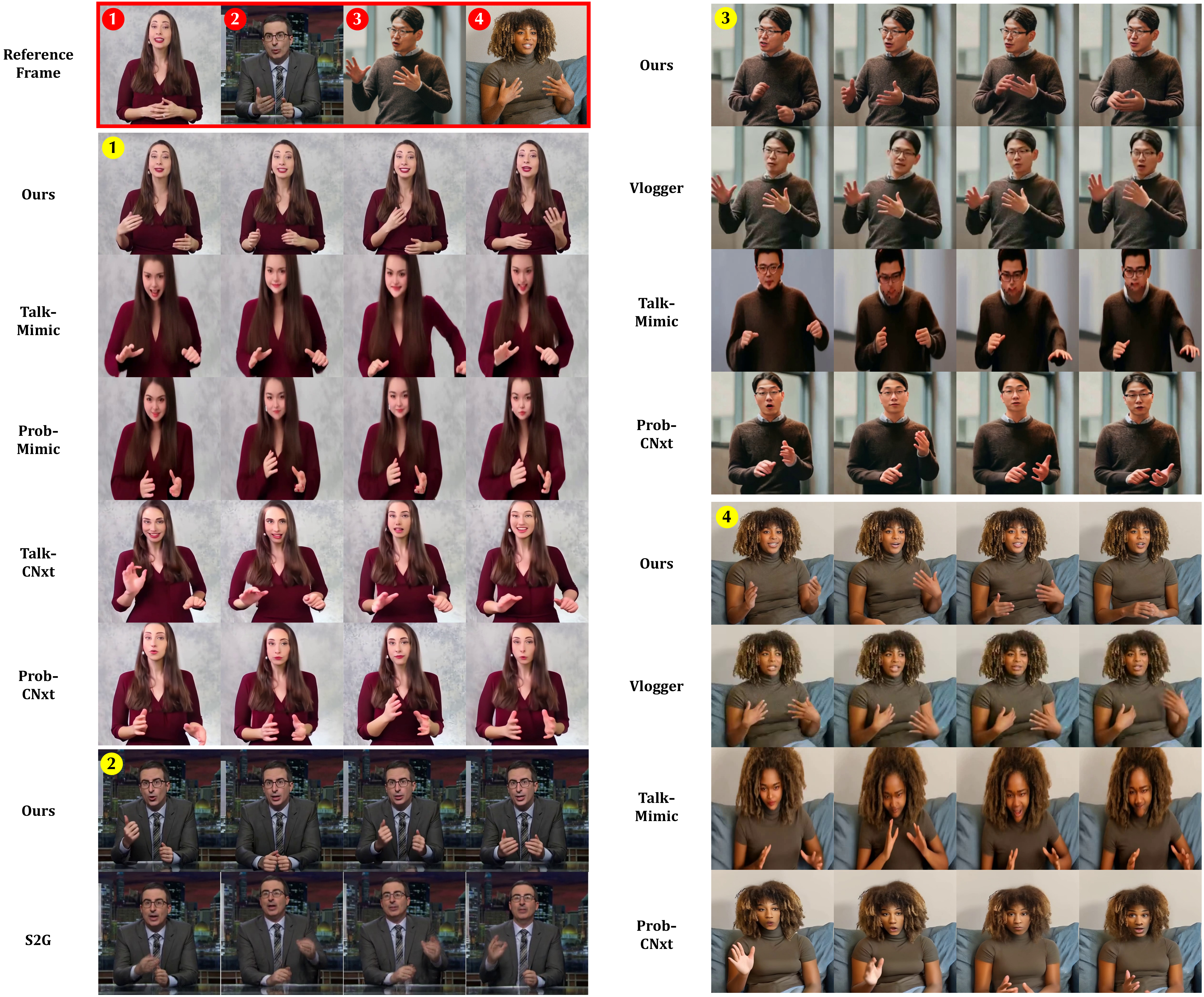}
\vspace{-15pt}
\caption{
\textbf{Qualitative Comparisons}. 
Frames used as reference are marked inside a red box, while frames generated conditioned on audio and the visual appearance of the corresponding reference frame are marked with numbers in yellow circles. 
Comparisons in this context should focus on factors, including image quality, the realism of poses, and the consistency of appearance relative to the reference frame.
}
\vspace{-10pt}
\label{fig:cmp_sota}
\end{figure*}

\section{Experiments}
\label{sec:experiment}
In this section, we begin by outlining the experimental setup, including implementation details and the baselines for comparison. Next, we 
comprehensively present both quantitative and qualitative results to showcase the performance of our method. Finally, we perform an ablation study to assess each component in our designs.

\noindent\textbf{Implementation Details.}
Our model is built based on success in general video generation, with 30 stacked MM-DiT-like Transformer layers.
We initialize parts of our model with pretrained weights from the CogVideoX-2B~\cite{yang2024cogvideox}.
Our model training is performed on a carefully curated dataset of about 300 hours of videos, specifically collected to feature speaking humans positioned centrally within the frame. 
The training process spans about 14 days on 8 A100 GPUs using a learning rate of $5e^-5$.
We evaluate our method under a zero-shot protocol, meaning that the test inputs, including driving audio and reference frames, are entirely unseen during training, and there is no overlap in the subjects between training and testing data. 
The test set consists of a diverse selection of video clips and frame-audio pairs: 50 clips from 10 subjects in videos collected by us, 
20 clips from 4 subjects sourced from the PATs~\cite{ahuja2020style}, 
and 10 frame-audio pairs collected from the homepage of Vlogger~\cite{corona2024vlogger} (Vlogger demo set). 
Each video or audio clip lasts $3\sim10$ seconds.

\noindent\textbf{Comparison Methods.}
Generalized audio-driven human video generation is rarely explored in existing research, so we primarily combine methods from two aspects as baselines. 
For audio-driven human gesture generation, we employ methods including TALKSHOW~\cite{yi2022generating} and ProbTalk~\cite{liu2024towards}, which synthesize human gestures based on audio inputs. The generated outputs are then converted from a 3D parametric format to human skeletons, aligning with the driving representation used in pose-driven human video reenactment approaches.
In the second aspect, we incorporate two open-source models: ControlNext~\cite{peng2024controlnext} and MimicMotion~\cite{zhang2024mimicmotion}. Both are trained on extensive human videos, enabling zero-shot reenactment.
Based on these models, we have four combined baselines, enabling generalized audio-driven human video generation for our comparisons. We denote them as \textbf{Prob-CNxt}, \textbf{Prob-Mimic}, \textbf{Talk-CNxt}, \textbf{Talk-Mimic}, repsectively.
Comparisons with \textbf{Vlogger}~\cite{corona2024vlogger} are also demonstrated in our experiments. We download their results with reference images and driving audios from their homepage for fair comparisons.

We also include a comparison with a co-speech generation method \textbf{S2G}~\cite{he2024co}, which is optimized for capturing the motion patterns of a specific subject. Using the official implementation of their model, we conduct evaluation on the subject Oliver~\cite{ahuja2020style}.
It is important to note that our method remains in a zero-shot setting throughout this comparison, with no tuning specifically for this subject.

\subsection{Quantitative Comparisons}
\label{sec:Quantitative Comparisons}
We evaluate the generated human gesture videos across four aspects: 1) visual quality, 2) temporal coherence, 3) animation quality, and 4) identity fidelity.
For visual quality, we assess the static frames using \textbf{SSIM}~\cite{wang2004image}, \textbf{LPIPS}~\cite{zhang2018unreasonable}, and \textbf{FID}~\cite{heusel2017gans}, which capture the structural, perceptual, and distributional aspects of image quality.
For temporal coherence, we follow the approach in \cite{wang2023disco}, 
reporting \textbf{FID-VID}~\cite{balaji2019conditional} and \textbf{FVD}~\cite{unterthiner2018towards}.
To evaluate animation quality, we measure the landmark detection confidence in the body and hands regions using OpenPose~\cite{cao2017realtime}, which are denoted as \textbf{Body-C} and \textbf{Hand-C}, respectively. 
Additionally, we report the variance of hand landmarks to reflect the expressiveness and diversity of hand movements (denoted as \textbf{Hand-V}).
Following \cite{he2024co}, we also report the Beat Alignment Score (\textbf{BAS}) reflecting audio-gesture coherence. 
Finally, identity fidelity is assessed by calculating the cosine similarity (\textbf{CosSim}) between the face in the reference frame and each generated frame, leveraging a face recognition model. 

Experimental results are tabulated in Table~\ref{tab:cmp_sota_metrics}.
Compared with the co-speech generation method S2D, our method demonstrates significantly better outcomes in video quality. S2D obtains better performance in hand dynamics and identity preservation from person-specific training. Optimizing on a highly dynamic style of the subject Oliver~\cite{ahuja2020style} leads to a relatively large hand diversity. 
In contrast, our method provides more stable outputs, achieving better harmony between audio and gestures, as evidenced by a higher BAS score.
The results on the Vlogger demo set further validate the superior performance of our method.
In terms of animation quality, the two methods both demonstrate realistic bodies, and our method surpasses Vlogger in better diversity of hand movements.
In comparing the four combined baselines on a mixed test set of our collected videos and those from PATs, our generation framework achieves noticeably superior results. Although audio-driven gesture generation methods offer a wider variety of hand gestures, its generation lacks the understanding of the reference frame, resulting in movements that feel disjointed or misaligned. Furthermore, when these methods are cascaded with pose-driven approaches, error accumulation across the two stages compounds, leading to a decline in overall performance.

\subsection{Qualitative Results}
\label{sec:Qualitative Results}
Given that our task involves multi-modal generation within video sequences, evaluating solely on static frames would not provide a comprehensive assessment. We strongly encourage readers to view our supplementary video, which provides a clearer and more intuitive understanding of the generated results, especially considering the harmony between audio cues and motion dynamics. 

Here, we first present several samples compared with our baselines in Fig.~\ref{fig:cmp_sota}. 
In the first set of figures labeled with \textbf{\textcircled{1}}, we compare the results across four combined baselines. A key distinction is that our method maintains far superior appearance consistency with the provided reference frame. Among the first stage of baselines, TALKSHOW generates a wider range of gestures compared to ProbTalk, which in turn tends to produce more stable body movements. Both methods deliver plausible human gestures; however, when rendered into pixel space, they suffer from noticeable identity loss and visual artifacts, resulting in outcomes significantly inferior to ours in terms of both fidelity and clarity.
In the second set of figures, labeled with \textbf{\textcircled{2}}, we provide a comparison with a co-speech generation method. Although their approach is specifically trained on this subject, our method demonstrates comparable performance in generating rich gestures. Furthermore, our approach significantly outperforms in visual quality, particularly in capturing fine details of the face and hands.
On the right side of the figure, we present results generated using the Vlogger demo set. Compared to the combined baselines, both our method and Vlogger achieve significantly better outcomes. However, when directly comparing with Vlogger, our approach consistently demonstrates superior identity preservation. Anchoring on the reference frame, it is evident that the human gestures produced by our method are more varied and expressive than those generated by Vlogger.

\begin{figure}[]
\centering
\includegraphics[width=\linewidth]{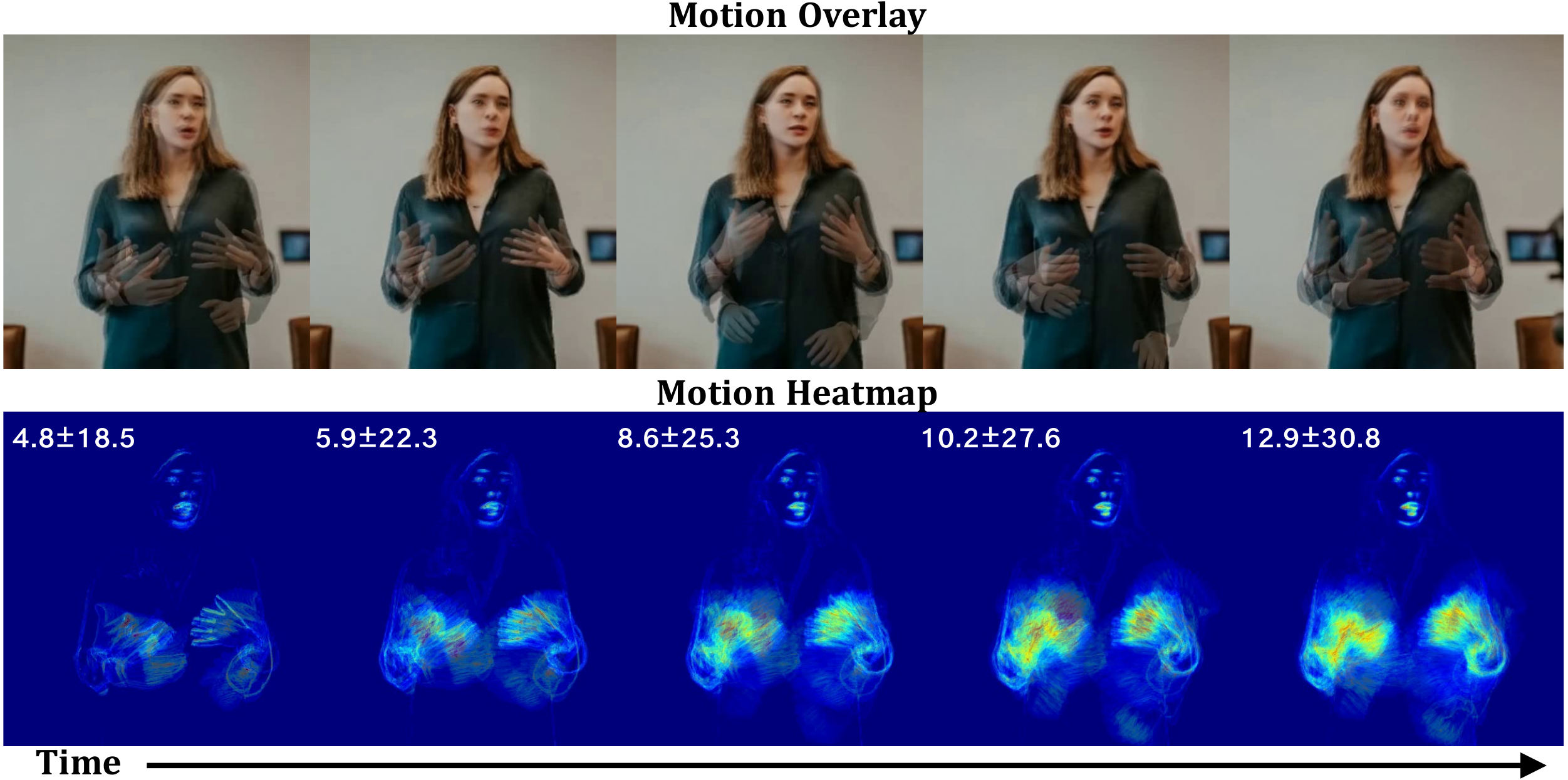}
\vspace{-15pt}
\caption{
The top row presents an overlay view of three generated videos. In the bottom row, we display the accumulated motion overlay, illustrating the overall movement patterns. The values in the top-left corner indicate the scaled mean and standard deviation of the motion, providing quantitative insight into the variability of the generated movements.
}
\vspace{-10pt}
\label{fig:diversity}
\end{figure}

\noindent\textbf{Animation Diversity}.
Our approach establishes a many-to-many mapping from audio signals to human motions. In Figure~\ref{fig:diversity}, we provide an intuitive visualization of the motion diversity achieved. We overlay three videos generated from the same reference-audio pair and present an accumulated motion heatmap, scaled within $[0,255]$. The mean and standard deviation values are displayed in the upper-left corner for reference. The static background demonstrates that our model accurately identifies relevant regions for audio-to-motion mapping. The variation in hand gestures across samples further validates the diversity of our method.

\subsection{Ablation Study}
\label{sec:ablation}
We evaluate our method against several modified versions of our model, described as follows:
1) ``\textbf{w/o R$^2$-DiT}'': This variant keeps only proposed H$^2$-DiT without leveraging the 3D prior for details refinery. It also serves as a baseline for comparison with all subsequent variants.
2) ``w/o Motion Token (\textbf{w/o MT})'': We exclude motion tokens from the input sequence, leaving only the noisy video tokens and audio tokens as inputs.
3) ``w/o Head Position Encoder (\textbf{w/o HPE})'': Here, the Head Position Encoder is removed.
4) ``w/o Appearance Adapter (\textbf{w/o AA})'': The Appearance Adapter within the Transformer layers is removed, allowing the model to access appearance information solely through prior frames.
5) ``w/o Identity Adapter (\textbf{w/o IA})'': This version removes the Identity Adapter.

\begin{figure}[]
\centering
\includegraphics[width=\linewidth]{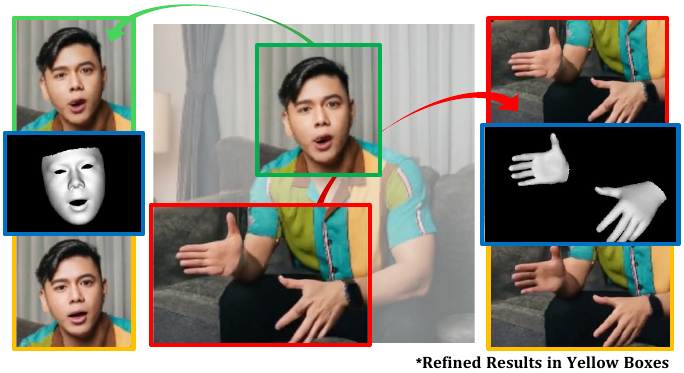}
\vspace{-15pt}
\caption{
Incorporating structural 3D priors enhances the capture of fine details, resulting in improved visual quality.
}
\vspace{-10pt}
\label{fig:cmp_3d}
\end{figure}

From the results in Table~\ref{tab:cmp_sota_metrics}, when the R$^2$-DiT is deprecated, the most distinct degradation is observed in hand confidence. Without 3D structural priors, the generated results demonstrate reasonable artifacts in details of the face and hands, as we give a qualitative comparison in Fig.~\ref{fig:cmp_3d}.
Removing the motion tokens does not impact a single forward pass; however, it significantly hinders the model’s ability to generate extended video sequences. Excluding the head position encoder leads to less stable video output, resulting in reduced detection confidence in the region of the body. Additionally, the two adapters integrated into our Transformer layers are crucial for accurately restoring details in the reference frame. Without these adapters, we observe a marked decline in overall visual quality and a substantial reduction in identity preservation.

%% file: sec/5_conclusion.tex
\section{Conclusion}

In this paper, we present a novel framework, {AudCast}, designed to tackle the challenging task of audio-driven human video generation. Our approach excels in producing high-quality human videos that feature synchronized lip movements and natural, rhythmic body motions aligned with audio input. Extensive comparisons with existing methods demonstrate the effectiveness of our cascaded DiT architecture, offering valuable insights and setting a foundation for future research in conditional human video generation under more intricate scenarios.

\noindent\textbf{Acknowledgments.}
This work is in part supported by National Natural Science Foundation of China with No. 62394322 and Beijing Natural Science Foundation with No. L222024.
This study is also supported by the Ministry of Education, Singapore, under its MOE AcRF Tier 2 (MOET2EP20221-0012, MOE-T2EP20223-0002), and under the RIE2020 Industry Alignment Fund – Industry Collaboration Projects (IAF-ICP) Funding Initiative, as well as cash and in-kind contribution from the industry partner(s).